\RequirePackage{fix-cm}
\documentclass[twocolumn]{svjour3}
\pdfoutput=1
\smartqed  
\usepackage{graphicx}
\usepackage{amsmath,amssymb}
\usepackage{url}
\usepackage[numbers]{natbib}
\usepackage[utf8]{inputenc}
\usepackage{mathptmx}      
\journalname{KI - K{\"{u}}nstliche Intelligenz}

\usepackage{microtype}

\begin{document}

\title{Neural Networks for Complex Data}

\author{Marie Cottrell  \and
        Madalina Olteanu \and
        Fabrice Rossi \and
        Joseph Rynkiewicz \and
        Nathalie Villa-Vialaneix
}


\institute{\'Equipe SAMM, EA 4543\\
           Universit\'e Paris I Panth\'eon-Sorbonne\\
           90, rue de Tolbiac\\
           75634 Paris cedex 13 France\\
           \email{Firstname.Lastname@univ-paris1.fr}
}

\date{Received: date / Accepted: date}

\maketitle

\begin{abstract}
Artificial neural networks are simple and efficient machine learning
tools. Defined originally in the traditional setting of simple vector data,
neural network models have evolved to address more and more difficulties of
complex real world problems, ranging from time evolving data to sophisticated
data structures such as graphs and functions. This paper summarizes advances on
those themes from the last decade, with a focus on results obtained by members
of the SAMM team of Universit\'e Paris 1. 
\end{abstract}

\section{Introduction}
In many real world applications of machine learning and related techniques,
the raw data are not anymore in a standard and simple tabular format in which
each object is described by a common and fixed set of numerical
attributes. This standard vector model, while useful and efficient, has some
obvious limitations: it is limited to numerical attributes, it cannot handle
objects with non uniform descriptions (e.g., situations in which some objects
have a richer description than others), relations between objects (e.g.,
persons involved in a social network), etc. 

In addition, it is quite common for real world applications to have some
dynamic aspect in the sense that the data under study are the results of a
temporal process. Then, the traditional hypothesis of statistical independence
between observations does not hold anymore: new hypothesis and theoretical
analysis are needed to justify the mathematical soundness of the machine
learning methods in this context. 

Artificial neural networks provide some of the most efficient techniques for
machine learning and data mining \cite{osowskiEtAl2004}. As other solutions,
they were mainly developed to handle vector data and analyzed theoretically in
the context of statistically independent observations. However, the last
decade has seen numerous efforts to overcome those two limitations
\cite{hammerJain04NonStandardData}. We survey in this article some of the
resulting solutions. We will focus our attention on the two major artificial
neural network models: the Multi-Layer Perceptron (MLP) and the
Self-Organizing Map (SOM).

\section{Multi-Layer Perceptrons}
The Multi-Layer Perceptron (MLP) is  one the most well known artificial
neural network model (see e.g., \cite{bishop_NNPR1995}). On a statistical
point of view, MLP can be considered as a parametric family of regression
functions. Technically, if the data set consists in vector observations in
$\mathbb{R}^p$, that is if each object is described by a vector
$x=\left(x_1,\cdots,x_p\right)^T$, the output of a one hidden layer perceptron
with $k$ hidden neurons is given by
\begin{equation}\label{equation:MLP}
F_{\theta}(x)=\beta+\sum_{i=1}^ka_i\psi\left(w_i^Tx+b_i\right), 
\end{equation}
where the $w_i$ are vectors of $\mathbb{R}^p$, and the $\beta$, $a_i$ and $b_i$ are
real numbers ($\theta$ denote the vector of all parameters obtained by
concatenating the $w_i$, $a_i$ and $b_i$). In this equation, $\psi$ is a
bounded transfer function which introduces some non linearity in
$F_{\theta}$. Given a set of training examples, that is $N$ pairs $(X_i,Y_i)$,
the learning process consists in minimizing over $\theta$ a distance between
$Y_i$ the target value and $F_\theta(X_i)$ the predicted value. Given an error
criterion (such as the mean squared error), an optimal value for $\theta$ is
determined by any optimization algorithm (such as quasi Newton methods see
e.g. \cite{boydVandenbergheCVBook2004}), leveraging the well know
backpropagation algorithm \cite{werbos74Thesis} which enables a fast
computation of the derivatives of $F$ with respect to $\theta$. The use of a
one hidden layer perceptron model is motivated by approximation results
such as \cite{hornik_stinchcombe_white_NN1989} and by learnability results
such as \cite{white1990NeuralNetworks} (in statistical community, learning is
called estimation and learnability consistency). 

\subsection{Model selection issues for MLP}
It is well known since the seminal paper of \cite{lapedes_Farber_1987} that
MLP are an efficient solution for modeling time series whenever the linear
model proves to be inadequate. The simplest approach consists in building a
non linear auto-regressive model: given a real valued time series
$\left(Y_t\right)_{t\in \mathbb{N}}$, one builds training pairs
$Z_t=(U_t,Y_t)$, where $U_t$ is a vector in $\mathbb{R}^p$ defined by
$U_t=\left(Y_{t-1},\cdots,Y_{t-p}\right))$. Then a MLP is used to learn the
mapping between the $U_t$ (the past of the time series in a time window of
length $p$) and $Y_t$ (the current value of the time series), as in any
regression problem.

In order to avoid overlearning and/or large
computation time, the question of selecting the correct number of neurons or,
more generally, the question of model selection arises immediately.  Standard
methods used by the neural-networks community are based on pruning: one trains
a possibly too large MLP and then removes useless neurons and/or connection
weights. Heuristic solutions include Optimal Brain Damage
\cite{leCunEtAl1990OBD} and Optimal Brain Surgeon \cite{hassibiEtal1993OBS},
but a statistically founded method, SSM (Statistical Stepwise Method), was
introduced by \cite{cottrellEtAl1995SSM}. The method relies on the
minimization of the Bayesian Information Criterion (BIC). Shortly after,
\cite{yao2000} and \cite{rynkiewiczEtAl2001} proved the consistency (almost
surely) of BIC in the case of MLPs with one hidden layer.  These results,
established for time series, allow to generalize the consistency
results in \cite{white1990NeuralNetworks} for the iid case.

The convergence properties of BIC may be generalized even further. A first
extension is given in \cite{rynkiewicz2008}. The noise is supposed to be
Gaussian and the transfer function $\psi$ is supposed to be bounded and three
times derivable. Then \cite{rynkiewicz2008} shows that under some mild
hypothesis, the maximum of the likelihood-ratio test statistic (LRTS)
converges toward the maximum of the square of a Gaussian process indexed by a
class of limit score functions. The theorem establishes the tightness of the
likelihood-ratio test statistic and, in particular, the consistency of
penalized likelihood criteria such as BIC. Some practical
applications of such methods can be found in \cite{mangeas1997}. The
hypothesis on the noise was relaxed in \cite{rynkiewicz2012}. The noise is 
no longer supposed to be Gaussian, but only to admit exponential
moments. Under this more general assumption, BIC criterion is still
consistent (in probability). 

On the basis of the theoretical results above, a practical procedure for MLP
identification is proposed. For a one hidden layer perceptron with $k$ hidden
units, we first introduce 
\[
T_n(k)=\min_{\theta}\left(E_n(\theta)+a_n(k,\theta)\right),
\]
where  $E_n(\theta)$ is the mean squared error of the MLP for parameter
$\theta$ and $a_n(k,\theta)$ is a penalty term. Then we proceed as follows:
\begin{enumerate}
\item Determination of the right number of hidden units.
\begin{enumerate}
\item begin with one hidden unit, compute $T_n(1)$, 
\item add one hidden unit if $T_n(k+1)\leq T_n(k)$, 
 \item if $T_n(k+1)>T_n(k)$ then stop and keep $k$ hidden units for the model.
\end{enumerate}
\item Prune the weights of the MLP using classical techniques like SSM \cite{cottrellEtAl1995SSM}.  
\end{enumerate} 
Note that the choice of the penalty term $a_n(k,\theta)$ is very important. On
simulated data, good results have been reported for $a_n(k,\theta)$ from
$a_n(k,\theta)=\frac{E_n(\theta)\log(n)}{n}$ to
$a_n(k,\theta)=\frac{E_n(\theta)\sqrt(n)}{n}$ (see \cite{rynkiewiczEtAl2001},
\cite{rynkiewicz2006}).

Let us also mention that the tightness of the LRTS and, in particular, the
consistency of the BIC criterion were recently established for more complex
neural-networks models such as mixtures of MLPs \cite{olteanuRynkiewicz2008}
and mixtures of experts \cite{olteanuRynkiewicz2011}.

\subsection{Modeling and forecasting nonstationary time series}
As mentioned in the previous section, MLP are a useful tool for modeling time
series. However, most of the results cited above are available for iid data or
for stationary time series.  In order to deal with highly nonlinear or
nonstationary time series, a hybrid model involving hidden Markov models (HMM)
and multilayer perceptrons (MLP hereafter) was proposed in
\cite{rynkiewicz1999}. Let us consider $\left(X_t\right)_{t\in \mathbb{N}}$ a
homogeneous Markov chain valued in a finite state-space
$\mathbb{E}=\left\{e_1,\cdots,e_N\right\}$ and $\left(Y_t\right)_{t\in
  \mathbb{N}}$ the observed time series. The hybrid HMM/MLP model can be
written as follows:
\begin{equation}
 Y_{t+1}=F_{X_{t+1}}\left(Y_t,\cdots,Y_{t-p+1}\right)+\sigma_{X_{t+1}}\epsilon_{t+1},
\end{equation}
where $F_{X_{t+1}}\in\left\{F_{e_1},\cdots,F_{e_N}\right\}$ is a regression
function of order $p$. In this case, $F_{e_i}$ is the $i$-th MLP of the model,
parameterized by the weight vector
$w_i$. $\sigma_{X_{t+1}}\in\left\{\sigma_{e_1},\cdots,\sigma_{e_N}\right\}$ is
a strictly positive number and $\left(\epsilon_t\right)_{t\in \mathbb{N}}$ is
a iid sequence of standard Gaussian variables.

The estimating procedure as well as the statistical properties of the
parameter estimates were established in \cite{rynkiewicz2001}. The proposed
model was successfully applied in modeling difficult data sets such as ozone
peaks \cite{dutotEtAl2007} or financial shocks \cite{mailletEtAl2004}.

\subsection{Functional data}\label{subsection:fda}
The original MLP model is limited to vector data for an obvious reason: each
neuron computes its output as a non linear transformation $\psi$ applied to a
(shifted) inner product $w^Tx+b$ (see equation
\eqref{equation:MLP}). However, as first pointed out in
\cite{sandberg_IEEETCS1996}, this general formula applies to any data space on
which linear forms can be defined: give a data space $\mathcal{X}$ and a set
of linear functions $\mathcal{W}$ from $\mathcal{X}$ to $\mathbb{R}$, one can
define a general neuron with the help of $w\in \mathcal{W}$, as calculating
$\psi(w(x)+b)$. 

This generalization is particularly suitable for functional data, that is for
data in which each object is described by one or several functions
\cite{ramsay_silverman_FDA1997}. This type of data is quite common for
instance in multiple time series setting (where each object under study
evolves through time and is described by the temporal evolutions of its
characteristics) or in spectrometry. A functional neuron
\cite{rossi_conanguez_NN2005} can then be defined as calculating
$\psi\left(b+\int fw\, \mathrm{d}\mu\right)$, where $f$ is the observed
function and $w$ is a parameter function. Results in
\cite{rossi_conanguez_NN2005} show that MLP based on this type of neurons
share many of the interesting properties of classical MLP, from the universal
approximation to statistical consistency (see also
\cite{rossi_conanguez_NPL2006} for an alternative functional neuron with
similar properties). In addition, the parameter functions $w$ can be
represented by standard numerical MLP, leading to a hierarchical solution in
which a top level MLP for functional data is obtained by using a numerical MLP
in each of its functional neurons. Experimental results in
\cite{rossi_conanguez_NN2005,rossi_conanguez_fleuret_IJCNN2002} show the
practical relevance of this technique.

\section{Self-Organizing Maps}
As the MLP, Kohonen's Self-Organizing Map (SOM) is one of the most well known
artificial neural network model \cite{kohonen_SOM2001}. The SOM is a
clustering and visualization model in which a set of vector observations in
$\mathbb{R}^p$ is mapped to set of $M$ neurons organized in a low dimensional
prior structure, mainly a two dimensional grid or a one dimensional
string. Each neuron $c$ is associated to a codebook vector $p_c$ in
$\mathbb{R}^p$ ($p_c$ is also called a prototype). As in all prototype based
clustering methods, each $p_c$ represents the data points that have been
assigned to the corresponding neuron, in the sense that $p_c$ is close to
those points (according to the Euclidean distance in $\mathbb{R}^p$). The
distinctive feature of the SOM is that each prototype $p_c$ is also somewhat
representative of data points assigned to other neurons, based on the
geometry of the prior structure: if neurons $c$ and $d$ are neighbours in the
prior structure, then $p_c$ will be close to data points assigned to neuron
$d$ (and vice versa). On the contrary, if $c$ and $d$ are far away from each
other in the prior structure, the data points assigned to one neuron will not
influence the prototype of the other neuron. This has some very important
consequences in terms of visualization capabilities, as illustrated in
\cite{vesanto1999SomVisu} for instance. 

The original SOM algorithm has been designed for vector data, but numerous 
adaptations to more complex data have been proposed. We survey here three
specific extensions, respectively to time series, functional data and
categorical data. Another important extension not covered here is proposed in
\cite{hammer_etal_N2004} which is built upon processing of multiple
time series with recursive versions of the SOM. The authors show that trees
and graphs can be clustered by those versions of the SOM, using a temporal
coding of the structure. Recent advances in this line of research include
e.g. \cite{hagenbuchnerEtAl2009}. Other specific adaptation include the symbol
strings SOM described in \cite{somervuo_NN2004}. 

\subsection{Time series with metadata}
While the SOM is a clustering algorithm, it has been used frequently in
supervised context as a component of a complex model. We described briefly
here one such model as an example of complex time series processing with the
SOM. Let us consider a time series with two time scales, i.e., that can be
written down with two subscripts. The date is denoted by $(j,h)$ where $j$
represents the slow time scale and corresponds for instance to the day (or
month or year) while $h=1, \ldots,H$ corresponds to the observed values
(e.g. the hours or half-hours of the day, the days of the month, the months of
the year, etc.). Then the time series is denoted $(c_{j})_{j\geq
  0}=\left((c_{j,1},\ldots,c_{j,H})\right)_{j\geq 0}$. We assume in addition
that the slow time scale is associated with metadata. For instance, if each
$j$ corresponds to a day in a year and one knows the day of the
week, the month, etc. Metadata are supposed to be available prior a
prediction.

The original time series $c_{j,h}$ takes value in $\mathbb{R}$, but the dual
time scale leads naturally to a vector valued time series representation, that
is to the $c_j\in \mathbb{R}^H$. In this point of view, given the past of the
vector valued time series, one has to predict a future vector value, that is a
complete vector of $H$ values. This could be seen as a long term forecasting
problem for which a usual solution would be to iterate one-step ahead
forecasts. However, this leads generally to unsatisfactory solutions either
because of a squashing behaviour (convergence of the forecasting to the mean
value of series) or to a chaotic behaviour (for nonlinear methods).

An alternative solution is explored in \cite{cott_1998}. It consists in
forecasting separately, on the one hand, the mean and variance of the time
series on next slow time scale step (that is, on the next $j$), and on the
other hand, the \emph{profile} of the fast time scale. The prediction of the
mean and of the variance is done by any classical technique. For the profile,
a SOM is used as follows. The vector values of the time series, i.e., the
$(c_{j})_{j\geq 0}$, are centred and normalized with respect to the fast time
scale, that is are transformed into profiles defined by
\begin{equation}\label{equation:profile}
q_j=\frac{1}{\sigma_j}\left((c_{j,1}-\mu_j,\ldots,c_{j,H}-\mu_j)\right),
\end{equation}
where $\mu_j=\frac{1}{H}\sum_{h=1}^Hc_{j,h}$ and
$\sigma^2_j=\frac{1}{H}\sum_{h=1}^H(c_{j,h}-\mu_j)^2$ are respectively the
mean and the variance of $c_j$. The profiles are clustered with a SOM leading
to some prototype profiles $p_c$. Each prototype is associated to the metadata
of the profiles that has been assigned to the corresponding neuron. 

Then a vector value is predicted as follows: the mean $\mu$ and variance
$\sigma$ are obtained by a standard forecasting model for the slow time
scale. Then the metadata of the vector to predict is matched against the
metadata associated to neurons: assume for instance, that metadata are days of
the week, and the we try to predict a Sunday. Then one collects all the
neurons to which Sunday profiles have been assigned. Finally, a weighted
average of the matching prototypes is computed and rescaled according to $\mu$
and $\sigma$. As shown in \cite{cott_1998} this technique enables both some
stable and meaningful full day predictions, while integrating non numerical
metadata.

\subsection{Functional data}
The dual time scale approach described in the previous section has become a
standard way of dealing with time series in a functional way, as shown in
e.g. \cite{besseCardotStephenson00}. But as pointed out in Section
\ref{subsection:fda}, functional data arise naturally in other contexts such
as spectrometry. Then, the SOM has been naturally adapted to functional data
in other contexts than time series. In those contexts, in addition to the
normalization technique described above that produces profiles, one can use
functional transformation such as derivative calculations in order to drive
the clustering process by the shapes of the functions rather than mainly by
their average values \cite{rossiConanGuezElGolliESANN2004SOMFunc}.

Another adaptation consists in integrating the SOM with optimal segmentation
techniques that represent functions or time series with simple models, such as
piecewise constant functions for instance. The main idea it to a apply a SOM
to functional data using any functional distance (from the $L^2$ norm to more
advanced Sobolev norms \cite{villmann2007Sobolev}) with an additional
constraint that prototypes must be simple, e.g., piecewise constant. This
leads to interesting visualization capabilities in which the complexity of the
display is automatically globally adjusted \cite{hebrailEtAl2010ClustSeg}.

\subsection{Categorical data}
In surveys, it is quite standard that the collected answers are categorical
variables with a finite number of possible values. In this case, a specific
adaptation of the SOM algorithm can be defined, in the same way that Multiple
Correspondence Analysis is related to Principal Component Analysis. More
precisely, useful encoding methods for categorical data are the Burt Table
(BT), which is the full contingency table between all pairs of categories of
the variables, or the Complete Disjunctive Table (CDT), that contains the
answers of each individual coded as 0/1 against dummy variables that
correspond to all the categories of all variables. Then, a Multiple
Correspondence Analysis of the BT or of the CDT is nothing else than a
Principal Component Analysis on BT or CDT, previously transformed to take into
account a specific distance between the rows and a weighting of the
individuals \cite{leRoux:2004}. The SOM can be adapted to categorical data
using this approach, as described in \cite{cott_2005} and
\cite{cott_2004}. The same transformation on BT or CDT is achieved and a SOM
using the rows of the transformed tables can thus be trained. This training
provides an organized clustering of all the possible values of the categorical
variables on a prior structure such as a two dimensional grid. Moreover, if a
simultaneous representation of the individuals and of the values is needed,
two coupled SOM can be trained and superimposed. The aforementioned articles
present various real-world use cases from socio-economic field.

\section{Kernel and dissimilarity SOM}
The extensions of artificial neural networks model described in the previous
sections are \emph{ad hoc} in the sense that they are constructed using
specific features of the data at hand. This is a strength but also a
limitation as they are not universal: given a new data type, one has to design
a new adaptation of the general technique. In the present section, we present
more general versions of the SOM that are based on a dissimilarity or a kernel
on the input data. Assuming the existence of such a measure is far weaker than
assuming the data are in a vector format. For instance, it is simple to define
a dissimilarity/similarity between the vertices of a graph, a data structure
that is very frequent in real world problems \cite{newman2003GraphSurveySIAM},
while representing directly those vertices as vectors is generally difficult. 

\subsection{Dissimilarity SOM}\label{som-dissimilarity}
Let us assume that the data under study belong to a set $\mathcal{X}$ on which
a dissimilarity $d$ is defined: $d$ is a function from
$\mathcal{X}\times\mathcal{X}$ to $\mathbb{R}^+$ that maps a pair of objects
$x$ and $y$ to a non negative real number which measures how different $x$ and
$y$ are. Hypothesis on $d$ are minimal: it has to by symmetric
($d(x,y)=d(y,x)$) and such that $d(x,x)=0$.

As pointed out above, dissimilarities are readily available on sets of non
vector data. A classical example is the string edit distance
\cite{levenshtein1966} which defines a distance\footnote{A distance is a
  dissimilarity that satisfies in addition the strong hypothesis of the
  triangle inequality: $d(x,y)\leq d(x,z)+d(z,y)$.} on symbol strings. More
general edit distances can be defined, such as for instance the graph edit
distance which measure distances between graphs \cite{bunke_ICMLDMPR2003}. 

As the hypothesis on $\mathcal{X}$ are minimal, one cannot assume anymore that
vector calculation are possible in this set. Then, the learning rules of the
SOM do not apply as they are based on linear combination of the prototypes
with the data points. To circumvent this difficulty,
\cite{kohonen_somervuo_N1998} suggest to chose the
values of the prototypes $p_c$ in the set of observations $(X_i)_i$. This
leads to a batch version of the SOM which proceeds as follows. After a random
initialization of the prototypes, each observation is assigned to the neuron
with the closest propotype (according to the dissimilarity measure) and the
prototypes are then updated. For each neuron, the updated $p_c$ is chosen
among the observations as the minimizer the following distortion
\begin{equation}\label{eq:local:distortion}
	\sum_i \Gamma(N(X_i),c) d(X_i,p)
\end{equation}
where $N(X_i)$ is $X_i$'s neuron and $\Gamma$ is a decreasing function of the
distance between neurons in the prior structure. This modification of the SOM
algorithm is known as the \emph{median SOM} and is closely related to the
earlier median version of the standard k-means algorithm
\cite{kaufman_rousseeuw_STABLNRM1987}.

In the case where $(X_i)_i$ is a small sample, the constraint to chose the
prototypes in the data can be seen as too strong. Then,
\cite{elgolli_etal_RSA2006} suggests to associate several prototypes (a given
number $q$) to each neuron. A neuron is represented by a subset of size
$q$ from $(X_i)$ and the different steps of the SOM algorithm are modified
accordingly. A fast implementation is described in
\cite{conanguez_etal_NN2006}.

A successful application of the dissimilarity SOM on real world data concerns
school-to-work transitions. In \cite{massoniEtAl2009}, we were interested in
identifying career-path typologies, which is a challenging topic for the
economists working on the labor market. The data was issued from the
``Generation'98'' survey by the CEREQ. The data sample contained information
about 16040 young people having graduated in 1998 and monitored during 94
months after having left school. The labor-market statuses had nine
categories, from permanent contracts to unemployed and including military
service, inactivity or higher education.

The dissimilarity matrix was computed using optimal matching distances
\cite{abbottTsay2000}, which are currently the main stream in economy and
sociology. The most striking opposition appeared between the career-paths
leading to stable-employment situations and the ``chaotic'' ones. The stable
positions were mainly situated in the west region of the map. However, the
north and south regions were quite different: in the north-west region, the
access to a permanent contract (red) was achieved after a fixed-term contract
(orange), while the south-west classes were only subject to transitions
through military service (purple) or education (pink). The stability of the
career paths was getting worse as we moved to the east of the map. In the
north-east region, the initial fixed-term contract was getting longer until
becoming precarious, while the south-east region was characterized by the
excluding trajectories: unemployment (light blue) and inactivity (dark blue).

 \begin{figure}[htbp]
\begin{center}
  \includegraphics[height=0.6\linewidth,width=0.9\linewidth]{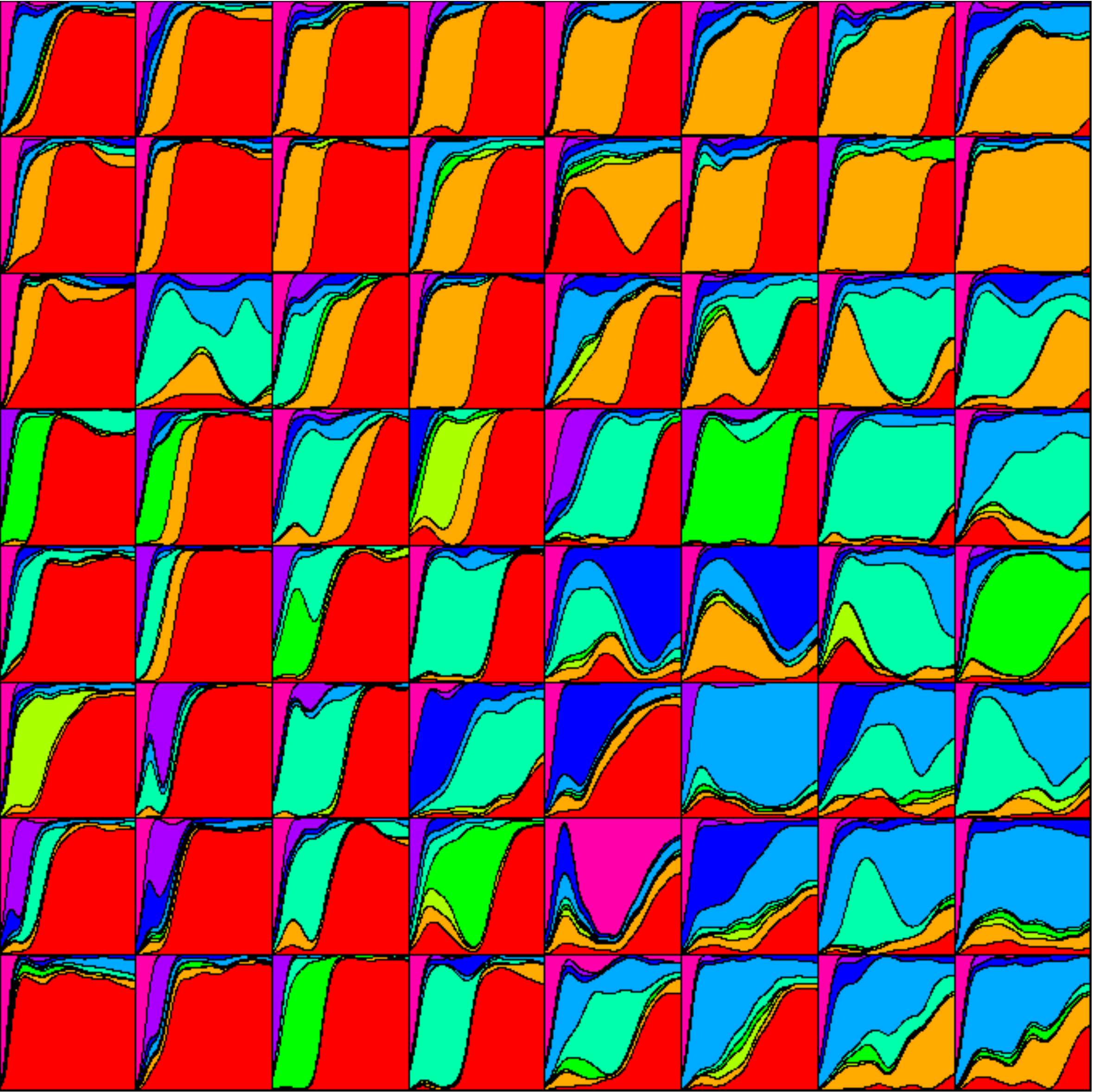}
\end{center}
  \caption{Career-path visualization with the dissimilarity SOM
    \cite{massoniEtAl2009}: colors correspond to the nine different categories}
\label{fig:career}
\end{figure}

Two other extensions of the SOM to dissimilarity data have been
proposed; they both avoid the use of constrained prototypes. The oldest one is
based on deterministic annealing \cite{graepel_etal_N1998} while a more recent
one uses the so-called relational approach that relies on pseudo-Euclidean
spaces \cite{hammerhasenfuss2010neuralcomputation,hammer_etal_WSOM2007}. Both
approaches lead to better results for datasets where the ratio between the
number of observations and the number of neurons is small. 

\subsection{Kernel SOM}\label{som-kernel}
An alternative approach to dissimilarities is to rely on kernels. Kernels can
be seen as a generalization of the notion of similarity. More precisely, a
kernel on a set $\mathcal{X}$ is a symmetric function $K$ from
$\mathcal{X}\times\mathcal{X}$ to $\mathbb{R}$ that satisfies a positivity
property:  
\begin{multline*}
\forall N\in\mathbb{N}^*,\ \forall (x_i)_{1\leq i\leq N}\in\mathcal{X}^N,\
\forall (\alpha_i)_{1\leq i\leq N}\in\mathbb{R}^N,\\
 \sum_{i,j=1}^N \alpha_i\alpha_j K(x_i,x_j)\geq 0.
\end{multline*}
For such a kernel, there is a Hilbert space $\mathcal{H}$ (called the feature
space of the kernel) and a mapping $\phi$ from $\mathcal{X}$, such that the
inner product in $\mathcal{H}$ corresponds to the kernel via the mapping, that
is \cite{aronszajn_TAMS1950} :
\begin{equation}
	\label{eq:kernel-trick}
	\langle \phi(x),\phi(x')\rangle_\mathcal{H} = K(x,x').
\end{equation}
Then $K$ can be interpreted as a similarity on $\mathcal{X}$ (values close to
zero correspond to unrelated objects) and defines indirectly a distance
between objects in $\mathcal{X}$ as follows:
\begin{align}\label{equation:kernel:distance}\nonumber
d_K(x,x')&=\|\phi(x)-\phi(x')\|_{\mathcal{H}}\\
&=\sqrt{K(x,x)+K(x',x')-2K(x,x')}.
\end{align}
As shown in e.g. \cite{shawetaylor_cristianini_KMPA2004}, kernels are a very
convenient way to extend standard machine learning methods to arbitrary
spaces. Indeed, the feature space $\mathcal{H}$ comes with the same elementary
operations as $\mathbb{R}^p$: linear combination, inner product, norm and
distance. Then, one has just to work in the feature space as if it were the
original data space. The only difficulty comes from the fact that $\phi$ and
$\mathcal{H}$ are not explicit in general, mainly because $\mathcal{H}$ is an
infinite dimensional functional space. Then one has to rely on equation
\eqref{eq:kernel-trick} to implement a machine learning algorithm in
$\mathcal{H}$ completely indirectly using only $K$. This is the so called
\emph{Kernel trick}.

In the case of the batch version of the SOM, this is quite simple
\cite{boulet_etal_N2008}. Indeed, assignments of data points to neurons are
based on the Euclidean distance in the classical numerical case: this
translates directly into the distance in the feature space, which is
calculated solely using the kernel (see equation
\eqref{equation:kernel:distance}). Prototypes update is performed as
weighted averages of all data points: weights are computed with the $\Gamma$
function introduced in equation \eqref{eq:local:distortion} as a proxy for the
prior structure. It can be shown that those weights, which are computed using
the assignments only, are sufficient to define the prototypes and that they
can be plugged into the distance calculation, without needing an explicit
calculation of $\phi$. Variants of this scheme, especially stochastic ones,
have been studied in \cite{andras_IJNS2002,macdonald_fyfe_ICKIESAT2000}. It
should also be noted that the relational approach mentioned in the previous
section \cite{hammerhasenfuss2010neuralcomputation,hammer_etal_WSOM2007} can
be seen a relaxed kernel SOM, that is an application of a similar algorithm in
situations where the function $K$ is not positive. 

While kernels are very convenient, the positivity conditions might seem very
strong at first. It is indeed much stronger than the conditions imposed to a
dissimilarity, for instance. Nevertheless, numerous kernels have been defined
on complex data \cite{gartner08:_kernel_struc_data}, ranging from kernels on
strings based on substrings \cite{lodhi_JMLR2002} to kernel between the
vertices of a graph such as the heat kernel
\cite{kondor_lafferty_ICML2002,smola_kondor_COLT2003} (see
\cite{boulet_etal_N2008} for a SOM based application of this kernel to a
medieval data set of notarial acts). Two graphs can also be compared via a
kernel based on random walks \cite{gartner_etal_ACCLT2003} or on subtrees
comparisons \cite{ramon_gartner_WMGTS2003}.

\section{Conclusion}
Present days data are becoming more and more complex, according to several
criteria: structure (from simple vector data to relational data mixing a network
structure with categorical and numerical descriptions), time evolution (from a
fixed snapshot of the data to ever changing dynamical data) and volume (from
small datasets with a handful of variables and one thousand of objects to
terabytes and more datasets). Adapting artificial neural networks to those new
data is a continuous challenge which can be solved only by mixing different
strategies as outlined in this paper: adding complexity to the models enable
to tackle non standard behavior (such as non-stationarity), theoretical
guarantees limit the risk of overfitting, new models can be tailor made for
some specific data structures such as graph or functions, while generic
kernel/dissimilarity models can handle almost any type of data. The ability to
combine all those strategies demonstrates once again the flexibility of the
artificial neural network paradigm. 


\small
\bibliographystyle{spbasic}
\bibliography{biblio}   

\end{document}